# Machine Translation for Accessible Multi-Language Text Analysis

Edward W. Chew[1], William D. Weisman[1], Jingying Huang[1], Seth Frey[1,*]

[1] Department of Communication, University of California Davis, Davis, CA, USA

**\* Correspondence:** Seth Frey (sethfrey@ucdavis.edu) 376 Kerr Hall, 1 Shields Dr., Davis, CA 95616, USA

# Machine Translation for Accessible Multi-Language Text Analysis

**Abstract**

English is the international standard of social research, but scholars are increasingly conscious of their responsibility to meet the need for scholarly insight into communication processes globally. This tension is as true in computational methods as any other area, with revolutionary advances in the tools for English language texts leaving most other languages far behind. In this paper, we aim to leverage those very advances to demonstrate that multi-language analysis is currently accessible to all computational scholars. We show that English-trained measures computed after translation to English have adequate-to-excellent accuracy compared to source-language measures computed on original texts. We show this for three major analytics—sentiment analysis, topic analysis, and word embeddings—over 16 languages, including Spanish, Chinese, Hindi, and Arabic. We validate this claim by comparing predictions on original language tweets and their backtranslations: double translations from their source language to English and back to the source language. Overall, our results suggest that Google Translate, a simple and widely accessible tool, is effective in preserving semantic content across languages and methods. Modern machine translation can thus help computational scholars make more inclusive and general claims about human communication.

**Introduction**

Humans communicate in thousands of languages, and yet a single language, English, attracts the bulk of communication research. This not only has the effect of depriving other languages of adequate attention, but depriving English-focused scholars of any sense of where the language stands relative to others. The general use of English-trained tools for English-focused analyses in the social science community is particularly notable given the ubiquity of multi-lingual data and the power of modern computational natural language processing. For example, social media researchers on Twitter typically begin with raw data that is highly multilingual, before filtering out all tweets except for English, or some other single language. With such practices scholars miss a tremendous opportunity to test the generalizability of social media-observed big data claims. But bringing standard text analysis tools to the level of training and refinement that English-trained tools receive is a forbidding prospect that few multi-lingual scholars have the training, resources, and language background to pursue. We propose a simple alternative approach that makes texts from over 100 languages accessible to the full variety of analyses that are typically available to only English-focused scholars. Specifically we demonstrate that modern machine translation has reached a level of refinement necessary to preserve sentiment, lexical topics, and semantic distance, making multi-language datasets legible to state of the art English-trained tools. By providing a validation of state-of-the-art machine translation, along with easily adaptable demonstration code, we aim to broaden the horizon of computational research and support Communication scholars in increasing the relevance and generality of their work.

Google Translate, the most popular, accurate, and accessible multilingual neural machine translation service, offers translations for over 133 languages (Caswell, 2022). In this paper, we demonstrate the efficacy of Google Translate in retaining sentiment valence across translations of large hand-coded and machine-coded Twitter datasets composed of tweets in 16 global non-English languages from four language families, being of Indo-European, Uralic, Semitic, and Sinitic origin. With our findings that Google Translate preserves the sentiment of tweets, as well as other dimensions of semantics, scholars may be emboldened to utilize Google Translate and other multilingual neural machine translation services to expand the generalizability of their research. In so doing, non-English languages can benefit from advanced English-trained natural language processing tools, and computational findings normally restricted to the English language can be expanded upon to broaden scholars' knowledge of global social phenomena.

Academics use Twitter datasets for a wide range of scholarship, including sentiment analysis (e.g. Gautam & Yadav, 2014), algorithmic training (e.g. Braithwaite et al., 2016), and even coronavirus disease 2019 detection (e.g. Gharavi et al., 2020). English-language corpora have been used to predict election results (Nausheen & Begum, 2018), analyze consumer preferences (Ahmed & Danti, 2016), and explore pro- and anti-childhood vaccine communities' influence on Twitter (Featherstone et al., 2020).

As valuable as this work is, it can only be more valuable extended across languages. Frey et al. (2018) use a corpus of six languages to document general ripple effects of emotional influence through others and back around to the self.  Mocanu et al. (2013) use data on 78 languages to characterize inter-linguistic diversity and intra-linguistic drift across municipalities and regions. And Alshaabi et al. (2021) compare the dynamics of social influence on Twitter over 150 languages. In other disciplines, large-scale multi-language comparisons in other disciplines have identified universal patterns in the cross-language naming of colors (Lindsey & Brown, 2009), a well as a universal preference for the shortening of dependency length in human sentence production (Futrell et al., 2015).

Our research demonstrates the effectiveness of Google Translate on the maintenance of sentiments, topic clusters, and semantic distance for tweets in all languages we examine. We validate the approach using "backtranslation," a classic validation method of machine translation in which scholars compare an original text to a version of that text that has been translated from its original language to another language (in our case English) and then back again to the original (Figure 1). This makes it possible to directly compare the accuracy of English-trained tools on English translations to original-language-trained tools on original-language texts, while controlling for semantic drift introduced by the translation process itself. We first test the preservation of sentiments using two large public multi-lingual Twitter datasets, one with hand-coded sentiments (Mozetič et al., 2016) and another with machine-coded sentiments (Imran et al., 2022) for this analysis. The second portion of our research applies the same two datasets to show that Google Translate preserves topic clusters after backtranslation, thereby demonstrating a similar level of semantic conservation for this second common text analysis task. In the third and final portion of our present study we demonstrate the effectiveness of out-of-the-box machine translation on a third common text analysis approach: neural word embeddings. Here, after backtranslation, 10 of 16 languages examined performed better than chance in maintaining a minimal embedding distance. Our findings provide strong support for the use of modern machine translation for expanding academic attention to the languages spoken by most humans.

**Method**

**Datasets**

We utilized two large, multilingual Twitter datasets. First, we examine the Mozetič et al. (2016) dataset, which contains over 1.6 million general Twitter posts hand-labeled as containing "positive", "negative", or "neutral" labels for 15 European languages: Albanian, Bosnian, Bulgarian, Croatian, English, German, Hungarian, Polish, Portuguese, Russian, Serbian, Slovak, Slovenian, Spanish, and Swedish. To expand the scope of our research beyond European languages, we added tweets from the Imran et al. (2022) COVID-19 dataset, a larger (70 million tweet) corpus including tweets in Chinese, Hindi, and Arabic. While these two datasets are comparable (both include sentiment labels), they differ in subject and date, as well as in how they determined sentiment scores. Those in Mozetič et al. (2016) were applied by human language-domain experts, while tweets from Imran et al. (2022) dataset were determined by algorithms (all trained within-language).

**Data cleaning**

Before translation and subsequent analysis, we preprocessed all Twitter data to remove Twitter handles, return handles, URLs, numbers, empty tweets, and converting all content to lowercase. We dropped Serbian from our analysis halfway through the study, after discovering that the Mozetič dataset contains Cyrillic Serbian, but Google Translate only supports Latin-character Serbian. We obviously excluded all English language tweets from validation by backtranslation through English.

To reduce our dataset to a more manageable, and affordable, size (the Google Translate API is paid), we randomly sampled 30,000 tweets from each of the 13 applicable

European languages from Mozetič et al. (2016) dataset, and 10,000 tweets from Chinese, Hindi, and Arabic from the Imran et al. (2022) dataset, for a total of 16 languages.

**Translation process**

Utilizing the Google Translate API, we translate all tweets from their "original language" datasets into English, saving the results as our "English translated" dataset. We then translate all the English translated tweets back to their original language, saving it as our "backtranslated" dataset (Figure 1). Our results are based only partly on three-way comparisons between these datasets. Where there is not a meaningful correspondence between English- and original-language analyses we use only two-way comparisons between the original and backtranslated datasets.

With this manuscript we share the scripts and instructions, to enable researchers to easily extend their single-language corpus research to multiple languages. The code is available at https://osf.io/jx476/.

**Sentiment analysis**

We conduct our sentiment analysis task with the free open-source software *Polyglot* (Al-Rfou, 2015). *Polyglot* allows the generation of sentiment labels in more than 100 languages, with "-1" indicating negative sentiment, "0" indicating neutral sentiment, and "1" indicating positive sentiment for each word in each original-language tweet. Based on the difference between the number of positive sentiment words and negative sentiment words, we generate an overall polarity for each tweet. Polyglot's lexicon-based sentiment analysis relies on a valence dictionary of positive and negative words, computing the sentiment of a text as the simple sum of the valences of its words, normalized back down to the [-1, 1] interval. Our pipeline excluded neutrally labeled tweets: as a result of *Polyglot*'s lexicon-based sentiments, short texts like Twitter posts are overwhelmingly labeled as neutral which makes it difficult to distinguish the performance of sentiment analyses across translations.

We computed confidence intervals around the accuracy of each language's sentiments with bootstrapping (1000x). The final sentiment accuracies are the bootstrapped medians.

**Topic clustering**

While sentiment analysis is a common application for natural language tools, it only serves to answer a small range of questions. We expand our investigation of Google Translate's ability to preserve the content of translated text through topic analysis. Compared to sentiment analysis, topic analysis provides a more technical, but much more flexible approach to computationally representing the meanings of text. Although the process of topic analysis is language agnostic, common computational tools are typically built to only support the English language, from stopwords to supported character sets.

We model our topic clustering approach after Yin and Wang (2014) who present the open-source software GSDMM ("Gibbs Sampling algorithm of the Dirichlet Multinomial Mixture). GSDMM is trained upon short text posts such as those found within social media environments such as Twitter (Yin & Wang, 2014). We follow the original work's data cleaning steps of removing both emojis and stopwords. We excluded Albanian and Bosnian due to their incompatibility with our data cleaning dependencies.

Our cluster analysis process was as follows. For each language, we used a total of five iterations of the clustering algorithm. We then classified the backtranslated tweets to the clusters generated on the original language tweets. To estimate the success of machine translation at semantic preservation under topic analysis, we computed the proportion of backtranslated tweet that were correctly assigned to the cluster of their original-language version. We compare these proportions to the baseline "null" proportions expected by chance, as derived from random permutations of original cluster assignments. Like many clustering algorithms, GSDMM requires researchers to impose a desired number of clusters, rather than identifying the number of clusters through the same emergent process as cluster assignments. But the ability of backtranslation to preserve topic clusters depends on the number of clusters. Therefore, we observe the effectiveness of topic preservation across a range of clusterings by training models on each original language dataset for 2, 5, 10, 15, 20, 50, 100, 150, and 200 clusters.

Unlike with our evaluation of sentiment analysis, the analysis of topics is only able to compare original language and backtranslated analytics: it is not able to compare either to English. While the framework of sentiment analysis imposes the same meaning to the idea of "positive" and "negative" sentiments across languages, topics emerge from a narrow understanding of a word as the sequence of characters that constitutes it. To the extent that original language and backtranslated tweets use the same characters (as in languages borrowings from each other), they can be assigned to the same set of clusters. But lexicons in English and each original language are mostly non-overlapping, and there is ultimately no basis to map English translations to original-language-derived topics.

**Word embedding**

Polyglot (Al-Rfou, 2015) also supports semantic word embeddings across its languages. We determine semantic preservation under word embedding by embedding original and backtranslated tweets (as the normalized sum of the embeddings of their words) and calculating their (cosine) distance in the embedding's high-dimensional semantic space. Under this formalism, a distance of zero indicates perfect preservation of semantics after translation. Because Polyglot has a different semantic space for each language, it was not possible to compare the distance of the intermediate English texts to the original and backtranslated texts.

To measure how well machine translation preserves semantics under word embedding, we compared the embedding distances after backtranslation to two baseline distances. We computed the average *minimum* distance of tweets from 5,000 other tweets in that language and their average *average* distances. Our rationale is that meaning is preserved despite semantic drift imposed by the process of (double) machine translation if the average distance of backtranslated tweets from their originals is smaller than the average distances of different original language tweets from each other.

## Results

Our primary finding is that Google Translate is faithful enough to preserve the semantics of multilingual texts under three common text analysis tasks: sentiment analysis, topic analysis, and word embeddings.

**Application 1: Preservation of sentiment**

We find that overall accuracy of sentiment scores decreases less than 1% after backtranslation, from a median 65.17% accuracy (with a very tight 99% high-confidence interval (HCI) of [65.16, 65.18]) to 64.76% [64.75, 64.77]. While small, this decline was statistically significant, as measured by the separation of the 99% HCI bars. We display this result in Figure 2, below.

We did have one surprise from this process. We expected that the accuracy of English-trained sentiment on the English-translated tweets would be between or below the accuracy of the original or backtranslated tweets, whose "ground truth" sentiments were computed with models trained specifically for those languages. Instead median sentiment accuracy increases 4.95% following original languages' translation into English (original language median accuracy: 65.17%, HCI [65.16, 65.18]; English translated median accuracy: 70.12%, HCI [70.11, 70.13]). Somehow sentiments extracted from English translations are more accurate than sentiments of original language tweets, despite the process of translation in between (Figure 2). We speculate on this result in the Discussion section.

Looking specifically at how different languages performed, we found the expected decrease in accuracy rates between the original language datasets and the backtranslated datasets for Albanian, Arabic, Chinese, Slovak, and Spanish (Figure 3). Unexpectedly, the remaining language datasets, belonging to the languages of Bosnian, Bulgarian, Croatian, German, Hindi, Hungarian, Polish, Portuguese, Russian, Slovenian, and Swedish, experienced an increase in sentiment accuracy from original language to backtranslated form. Although languages on average showed higher accuracy in English translation, the original language datasets of Portuguese, Russian, Slovenian, and Swedish show a drop in sentiment accuracy when translated to English (while the remaining others, Albanian, Bosnian, Bulgarian, Croatian, English, German, Hungarian, Polish, Slovak, and Spanish all improve).

**Application 2: Preservation of lexical topic assignments**

Our primary finding from Application 2 is that machine translation also preserves topic structure (Figure 4). Specifically, backtranslated tweets are assigned to the same cluster and their original language version at a rate well above chance. As expected, the ability of backtranslation to preserve topic structure declines as the number of topics increases (Figure 5). Performance accuracy of topic cluster preservation is 78% when there are two topic clusters, and declines to about 60% when there are 10 or more clusters. However, chance accuracy declines much faster, from 52% to 10% over the same span. In other words, the relative accuracy of topic recovery actually *improves* with the number of clusters, even as absolute efficacy declines.

It may seem peculiar that chance performance of topic assignment remains at 10% even with 200 clusters. This is probably due to an unequal distribution of cluster sizes. For 200 equally sized clusters, the baseline, null probability of a backtranslated tweet being randomly assigned to its correct topic is 0.05%, one half of a percent. Chance performance higher than this is easy to arrange in a system with a few large clusters and a large number of very small ones.

Overall, we find that Google Translate preserves topic clusters across languages, with accuracy ranging from 60% to 80% depending on the number of topics we set the GSDMM algorithm to impose.

**Application 3: Preservation of semantics in embedding space**

In the final application of this work, we examine the multilingual preservation of semantic vectors in high-dimensional neural embeddings after machine translation and backtranslation.

On average, original language tweets are significantly closer to their backtranslations than to other original language tweets in the same collection (Figure 6). Across languages, average distances of original language tweets from each other are 0.041–0.184 units, their minimum distances from each other are 0.028–0.132, and their distances from their backtranslations are 0.203–0.492. Being less than half of the average distance (except Chinese) and below or slightly above the minimum distance, we can conclude that the semantic change introduced by the translation algorithm is enough to change the meaning of a backtranslated tweet to be mistakable for a different closely related tweet, but not the typical more distantly related tweet.

Of the 16 languages involved in the analysis, 6 languages (Albanian, Arabic, Chinese, German, Hindi, and Portuguese) failed the minimum baseline test, with backtranslated tweets having greater semantic distance from their originals than the average closest outside tweet. The Albanian, German, and Portuguese corpora failed by small margins (mean distance of 0.065 compared to minimum baseline distance 0.059 in Albanian; distance 0.110 compared to baseline 0.094 in German; 0.089 against 0.085 in Portuguese). But in Arabic, Chinese, and Hindi, embeddings of translations were even further from their original (distance 0.146 against 0.101 in Arabic; 0.184 against 0.053 in Chinese; 0.041 against 0.028 in Hindi). It should be noted that Arabic, Chinese, and Hindi were drawn from the Imran et al. (2022)

dataset focused on COVID-19 related tweets, included as part of our effort to expand this project's analysis beyond languages of European origin. As baseline measures were calculated on the distance between random tweets relative to their distance with all other tweets, and these tweets were semantically more closely related, these languages' baseline measures may have been especially narrow relative to those of the other languages as a result of their shared topic. Although they failed the rigorous minimum distance test, even Arabic, Chinese, and Hindi passed the mean distance test: they were closer in meaning to their original than the average tweet (mean baseline distances 0.416, 0.352, and 0.203, respectively).

## Discussion

As the global academic world becomes increasingly interconnected, Communication scholars must meet the challenge to make claims about communication processes more globally relevant. Fortunately, with recent advances in natural language processing, quantitative Communication research has an opportunity to be multilingual by default. Advances that bring equal attention to more of the world's languages will not only provide greater generality of results, but greater attention to the work of Communication scholars from all parts of the world. Standard approaches to large multilingual corpora will also allow the rapid transfer of groundbreaking knowledge to and from the international Communication community.

Of course, these advances have downsides. When multi-language analyses are conducted by scholars who can't speak all of those languages, it becomes harder for them to "gut check" or "sanity check" their results, culturally contextualize those results, and interpret whatever valid cross-linguistic differences that do appear,. By encouraging researchers to conduct multilingual studies by default, we are almost necessarily advocating for a circumstance in which scholars are making conclusions about languages that they do not know. Although this approach has some acceptance in other fields, such as large-scale comparative linguistics, it would be understandable to see it as controversial. As novel as this problem may be, the way forward may not be novel at all. Quantitative and qualitative methods have a fundamental complementarity, with the former bringing generality as the latter brings depth and sensitivity to context. By supporting the summary quantitative claims of non-speakers with citations to other work by native speakers and other domain experts, scholars may be able to justify not knowing the languages they are engaging with. This complementary approach will be particularly valuable for understanding outliers. In the case of our research, Chinese, Arabic, and Hindi all perform worse than the other languages. Having ruled out explanations that go to the phylogeny, character set, and geography of these languages, domain experts become the best candidates for understanding how and why specific languages deviate from the majority of their peers. This illustrates the importance of framing our contribution as a complement to expert-based multi-language communication research, rather than a substitute.

In this work we are able to validate the performance of Google Translate by leveraging source-language versions of our three methods: sentiment analysis, topic analysis, and word embeddings. However, as others make use of machine translation, they will not have the comfort of source-language tools, and may feel that they are "flying blind". Although we succeed in showing that translation introduces negligible drift, it may still be uncomfortable to apply it to a new dataset, particularly with text analysis methods beyond the three that we validate here (such as custom classifiers). To address this concern, researchers can take not only our conclusions, but the backtranslation method itself, to perform partial validations for their own case. Most likely, it should be possible to find "home language" tools for at least a handful of languages in a larger corpus. If an author can show satisfactory and stable performance across this subset, by comparing original and backtranslated texts, they can assure their audience that the method is probably working for other languages as well. Even lacking such "ground truth," there may be ways of using our method to instill confidence in a multilingual result. For example, a scholar could perform iterative backtranslations to calculate how many cycles must be introduced for the statistical significance of their result to degrade below threshold. If it takes a large number of backtranslations to degrade a result, readers can have confidence that artifacts introduced by the method are not sufficient to explain those results. Conversely, if machine translation is artificially amplifying a result, scholars can measure this effect with iterated backtranslation to suggest an appropriate amount of caution.

Another design choice of this work ensures the generality of the method we introduce. All three applications of this work were performed with tweets. Tweets are short, making them challenging for text analysis methods like sentiment and topic analysis. That our method is very effective on challenging text is encouraging for scholars who would extend this method to more typical (longer) texts.

A limitation of our approach is its accessibility. We have argued that Google Translate is very accessible, and this is true in that it requires a small amount of code (that we provide) to translate large quantities of text to English. However, our approach is not as financially accessible. The Google's Translation API costs $20 USD per million characters. In practice, this translates to roughly $100 USD per 130,000 tweets. Fortunately, free translation tools of comparable quality are increasingly common, and can also be validated in practice using backtranslation.

One surprising result from this work was that the accuracy of sentiment detection after translation into English, and in some cases after backtranslation, was higher than in the original texts. This finding is easier to understand with an appreciation of how sentiment analysis works in libraries like Polyglot. Polyglot uses the "dictionary" method, in which hundreds to thousands of high-frequency words are given sentiment scores, and the score of a statement is calculated from the sum of scores of the subset of words that are in the detector's sentiment dictionary. If the dictionary is large, or the text is long, then its assigned sentiment score will be based on a lot of signal. Consequently, this method is less suitable for rarer languages and shorter texts (like tweets), which are less likely to contain scored words. It is also more suitable to texts with more common words, since uncommon words are less likely to appear in a language's sentiment dictionary.

Why would translation to English, or backtranslation from English, improve task performance? Translation to English may be improving dictionary-based sentiment detection because English-language sentiment dictionaries tend to be longer. And subsequent backtranslation may be improving detection performance if it results in uncommon unscored words from the source text being backtranslated into more common words that are scored. We believe that this finding can be explained by Polyglots' apparent relative greater capacity to detect sentiment in English language content, relative to the content of other languages. This result underscores the need to validate Google Translate for each natural language task that it is being used to support.

**Conclusion**

There is an unmet need to extend Communication scholars' applications of text analysis to more languages, particularly in the data-rich context of social media studies. Translation tools such as Google Translate can be immensely helpful in meeting this need. We have quantified Google Translate's effectiveness in maintaining sentence meaning in translations to and from English. Across 16 non-English languages, sentiment analysis scores were shown to improve when translated to English, and only diminish marginally when translated back to their original languages. Similarly, both topic and semantic distances are preserved during backtranslation. Our findings demonstrate that machine translation is able to preserve semantic content and make non-English datasets legible to English-trained computational tools. We hope this analysis gives researchers the confidence to use machine translation to simply and economically increase the number of languages involved in their research, and thereby the generality of their findings.

## Acknowledgements

We would like to thank Arti Thakur, Communication Ph.D. Candidate at the University of California, Davis, for her assistance with the analysis. All code is available at [https://osf.io/jx476/](https://osf.io/jx476/). The authors report there are no competing interests to declare.

**Figure 1**

***We compare analytics computed on texts in their original languages to translated English language analytics and texts translated back to the original language.*** *Differences between the original and translated texts are typically difficult to attribute to semantic differences between the languages and "semantic" imposed by poor translation. Comparing original and backtranslated texts enables us to control for the effect of drift and focus on semantics.*

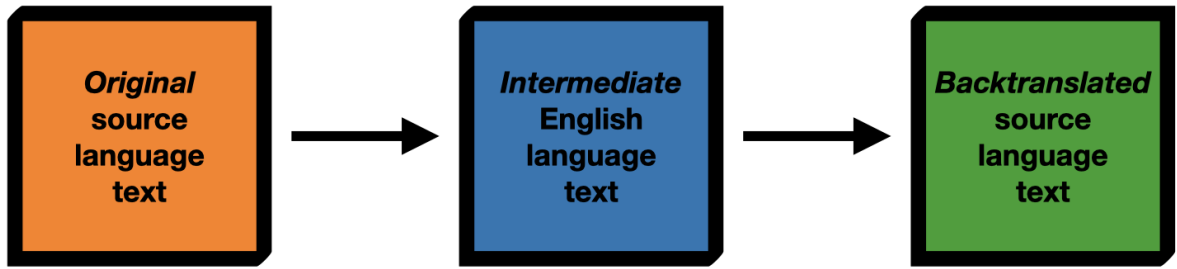

**Figure 2**

***Sentiment analysis overall retains accuracy after backtranslation by machine methods.***

*Median sentiment detection accuracy increases 4.9% from original language to English translated language datasets, and falls less than 1% from original language datasets to backtranslated language datasets. Note that 99% error bars are too narrow to be displayed.*

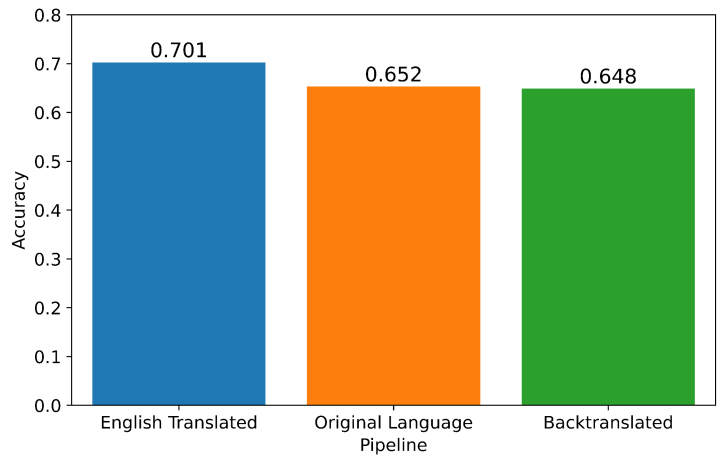

**Figure 3**

***Comparison of sentiment labeling accuracy across languages, before, during, and after backtranslation.*** *Seventeen language sentiment detection accuracy from original language > English translated > backtranslated datasets. Note that 99% error bars are too narrow to be displayed.*

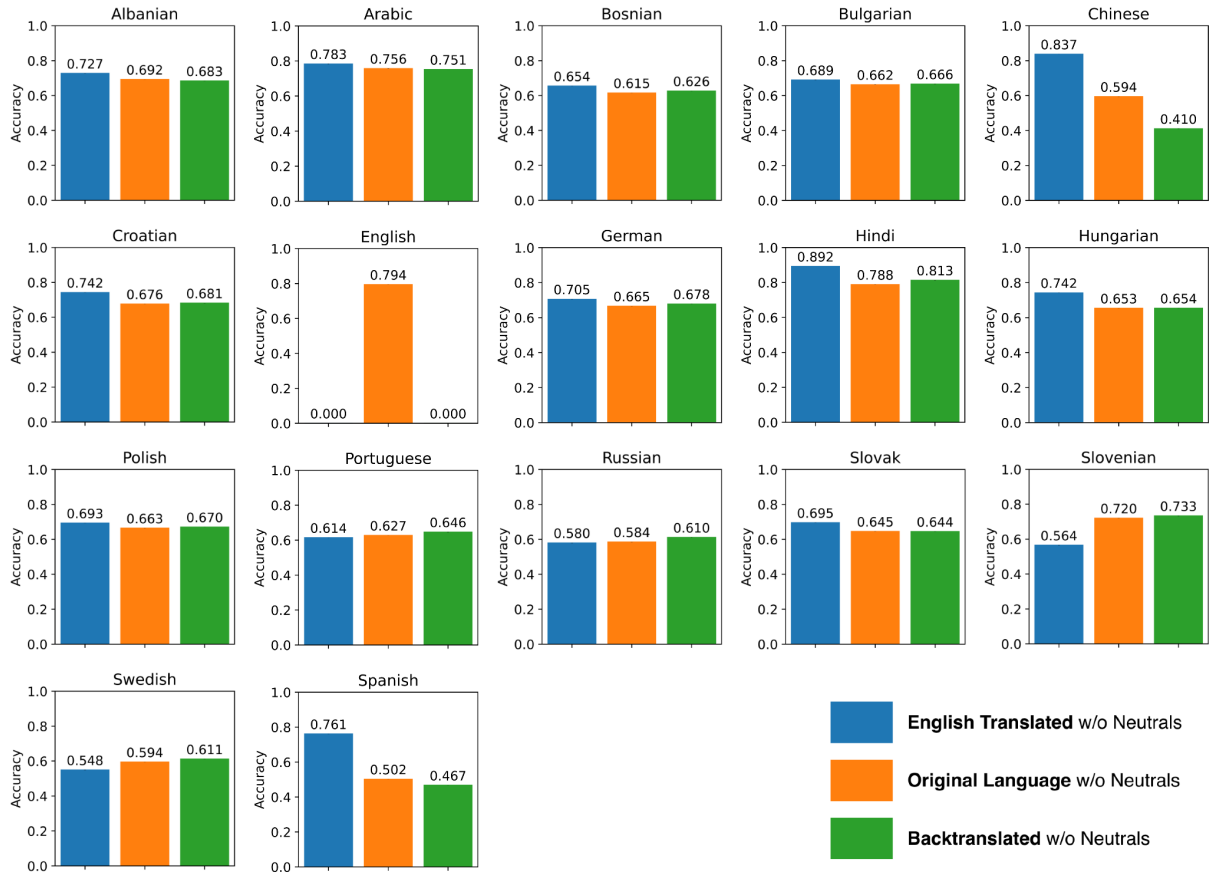

**Figure 4**

***Machine translation preserves topic clusters across languages, regardless of number of topics.*** *Percentage of backtranslated tweets assigned to the same cluster as original language tweets by language. White text denotes permutation test accuracy. The 14 languages examined demonstrate topic clustering accuracy preserved above chance.*

*2 Clusters*

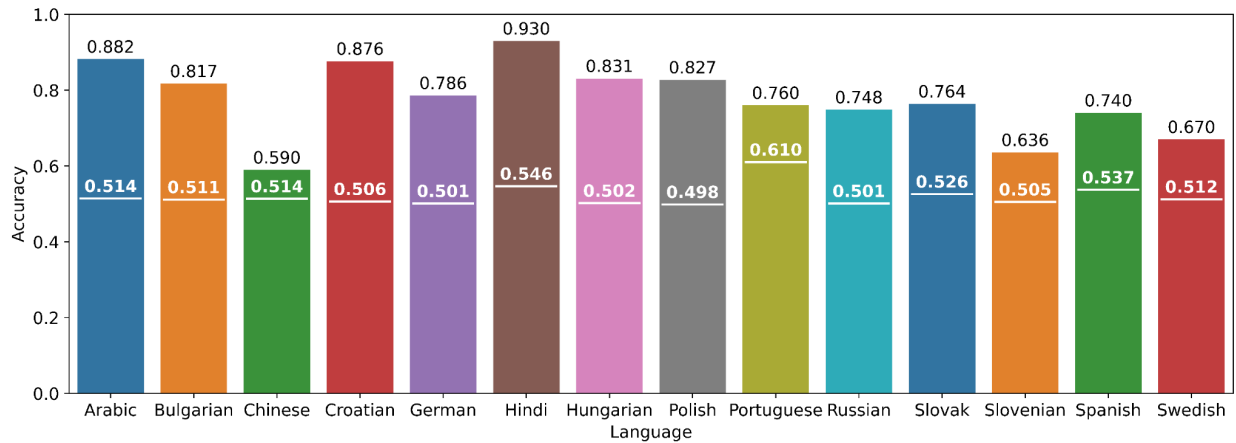

*100 Clusters*

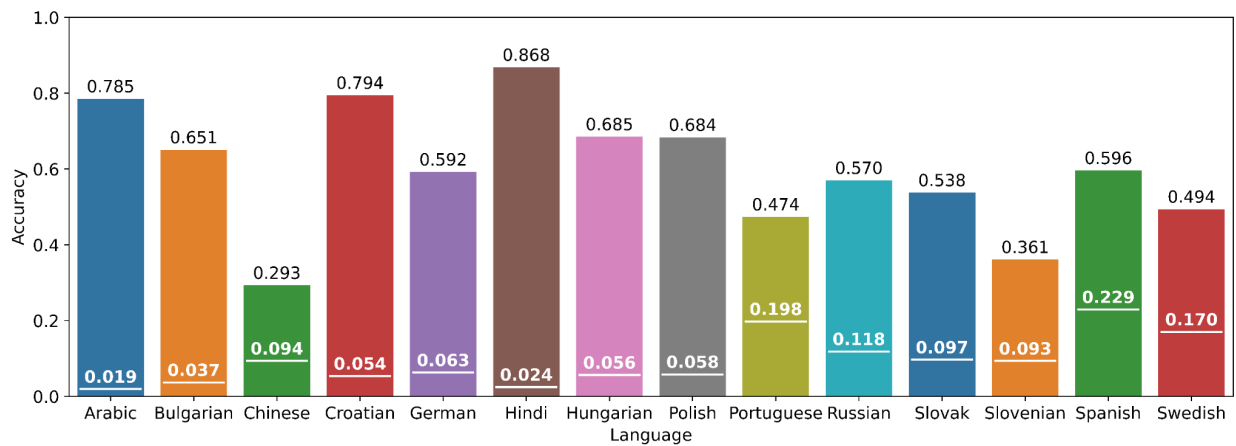

**Figure 5**

***Recovery of topics after backtranslation declines in an absolute sense with number of topics, but increases relative to baseline.*** *Average topic cluster accuracy by size of topic cluster. Note that topic cluster accuracy decreases from two to ten clusters, whereby it is approximately stable to 200 clusters.*

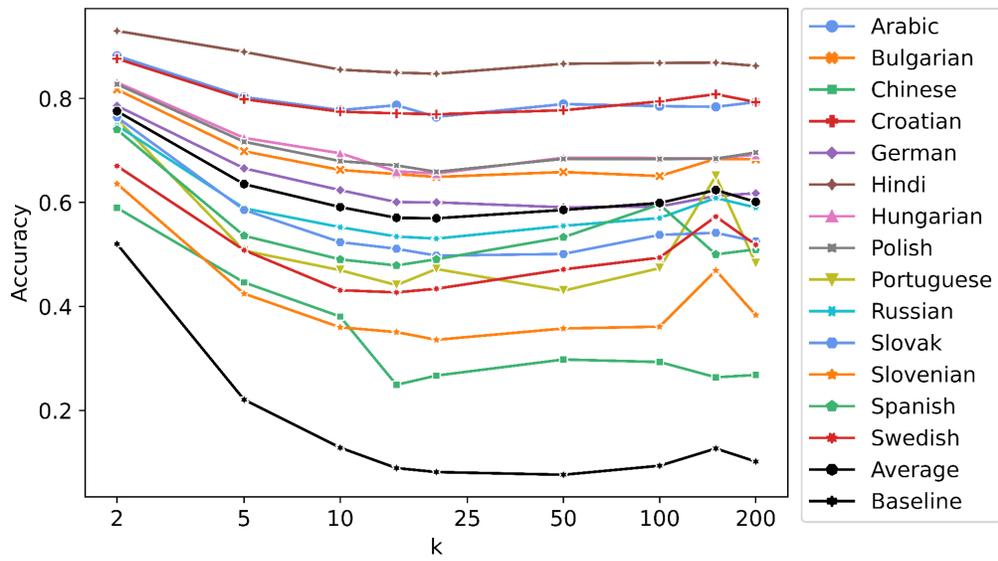

# Figure 6

***Tweets are closer to their backtranslations on average than to other tweets.***

*Average distance between original language and backtranslated sentence embeddings by language. Black lines denote the mean baseline distance and blue lines denote the minimum baseline distance. All 16 languages have mean distances below their mean baseline. All languages but Albanian, Arabic, Chinese, German, Hindi, and Portuguese have mean distances below their minimum baseline. In these languages, tweets backtranslated tweets are further from their source tweet in meaning than tweets that are very semantically similar to the source, but tweets in these languages are still consistently closer to their source than the average tweet.*

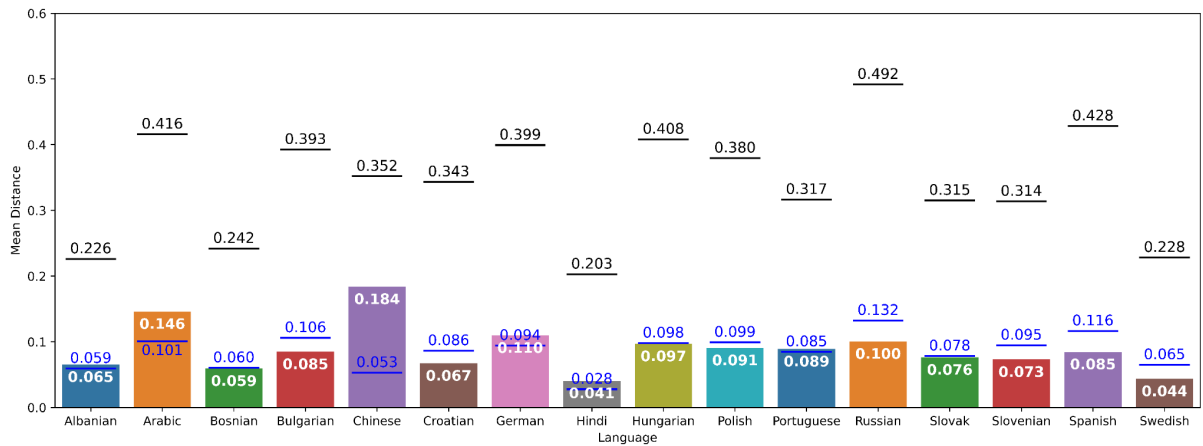